\documentclass[11pt,a4paper]{article}

\usepackage[margin=1in]{geometry}

\usepackage{amsmath,amssymb,amsthm}
\usepackage{mathrsfs}

\usepackage{graphicx}
\usepackage{float}
\usepackage{booktabs}
\usepackage{multirow}

\usepackage{xcolor}
\usepackage{authblk}
\usepackage{hyperref}

\usepackage[backend=biber,style=apa]{biblatex}
\DeclareLanguageMapping{english}{english-apa}
\hypersetup{
    colorlinks=true,
    linkcolor=blue,
    citecolor=blue,
    urlcolor=blue
}

\newcommand{\N}{\mathbb{N}}
\newcommand{\R}{\mathbb{R}}

\newcommand{\T}{\mathbb{T}}

\newcommand{\bx}{\mathbf{x}}
\newcommand{\bX}{\mathbf{X}}
\newcommand{\aX}{\mathbb{X}}
\newcommand{\bi}{\mathbf{i}}

\newcommand{\D}{\mathbb{D}}
\newcommand{\A}{\mathbb{A}}

\newcommand{\di}{\displaystyle}

\title{
\textbf{Signature-Based Feature Learning for Human Activity Recognition:\\
A Reproducible Machine Learning Study of Representation, Depth, and Model Choice}
}

\title{
\textbf{Signature-Based Feature Learning for Human Activity Recognition:\\
A Reproducible Machine Learning Study of Representation, Depth, and Model Choice}
}

\author[1]{Kamal Jarrar\thanks{Corresponding author: \href{mailto:jarrar.kamal@univ-pau.fr}{jarrar.kamal@univ-pau.fr}}}
\author[1]{Jacky Cresson}
\author[1]{Christian Paroissin}

\affil[1]{
Université de Pau et des Pays de l'Adour, CNRS, LMAP, Pau, France
}

\date{October 2026}

\begin{document}

\maketitle


\begin{abstract}
Human activity recognition (HAR) relies on transforming sensor signals into informative representations for classification. Although deep learning and handcrafted features are widely used, the role of representation itself is often not systematically isolated. Signature transforms provide a mathematically grounded way to encode temporal order and cross-channel interactions, but their value for HAR under a fully reproducible and leakage-aware framework remains unclear.

To evaluate whether signature-based feature learning improves HAR performance compared with raw-signal baselines, and to assess the effects of embedding strategy, truncation depth, model choice, and sensor configuration.

Experiments were conducted on the UCI HAR dataset using a fully reproducible pipeline with the original train--test split preserved and subject-disjoint validation to prevent leakage. Three representations were compared: raw flattened signals, time-augmented paths, and lead--lag transformed paths. Signature features were computed at multiple truncation depths and evaluated using multilayer perceptron (MLP) and Random Forest (RF) classifiers under identical preprocessing and validation procedures. A prior K-means-based feature reduction study was also reproduced for comparison.

Signature-based representations improved performance when paired with RF models, the best configuration was time-augmented six-channel signatures at depth 6 using entropy-based RF achieving 0.858 accuracy and 0.859 macro F1, outperforming the strongest raw baseline (0.816 accuracy). Lead--lag representations were competitive at moderate depths but did not surpass the best time-augmented models. MLP models did not exceed raw baselines.

Signature-based feature learning can improve HAR, but its benefit depends on alignment between representation design and classifier choice.
\end{abstract}


\vspace{1em}


\tableofcontents

\newpage


\section{Introduction}

\subsection{Human Activity Recognition (HAR)}

Human Activity Recognition (HAR) studies how to identify what a person is doing by using data collected from sensors. These activities include walking, sitting, and climbing stairs. In daily life, many devices such as smartphones and smartwatches contain small sensors that measure movement \parencite{lara2013survey,bulling2014tutorial}. These sensors, mainly accelerometers and gyroscopes, record how the body moves over time. Because these devices are widely available and do not capture images, they provide a low-cost and privacy-friendly way to monitor human behaviour \parencite{tran2023lowcost,climentperez2022privacy}. This is important in areas such as health monitoring, elderly care, and fitness tracking, where continuous observation is needed without disturbing the user \parencite{app132413009}.\\

In most HAR systems, the continuous stream of sensor data is divided into small segments called windows \parencite{anguita2013public,ucihar2012}. Each window contains a fixed number of time steps, often 128 values per channel. This design is used because it allows fast predictions, which are required in real-time applications \parencite{bulling2014tutorial,banos2014windowsize}. However, this choice creates a problem. A short window may not include a full movement cycle \parencite{xu2016wearable}. For example, a walking step may start in one window and end in another. Because of this, the model only sees part of the activity. This incomplete view makes it difficult to correctly distinguish between similar activities. As a result, the way we represent the data inside each window becomes very important.\\

Each data window can be written as a matrix that contains several time steps and multiple sensor channels \parencite{hammerla2016}. The key challenge is to transform this matrix into a form that clearly describes how the signal changes over time and how different channels interact with each other \parencite{hammerla2016}. Raw sensor values keep the original measurements, but they do not clearly show the order of movements or the relationships between channels \parencite{bulling2014tutorial}. These properties are important because human activities are defined by patterns that evolve over time, not just by individual values \parencite{bulling2014tutorial}. 

\subsection{Existing Approaches and Previous Work}

To the best of our knowledge, and following the survey by \textcite{wang2019deep}, two families of methods have been used to address the HAR problem, i.e., extracting {\bf features} from temporal data streams to identify the activities performed by a person.\\

{\bf Statistical methods}, which try to design features manually \parencite{lara2013survey}. For example, they compute averages, variances, or frequency components. These features can capture some patterns, but they may miss complex relationships between time steps \parencite{hammerla2016}.\\

{\bf Deep learning models}, which learn features automatically from raw data \parencite{9744439}. However, these models often receive flattened inputs, which means the temporal structure is not directly given to them \parencite{hammerla2016}. As a result, the model must learn both the representation and the classification simultaneously, which increases the complexity of the learning process and often reduces interpretability \parencite{bengio2013representation,lipton2018mythos}.\\

In this paper, we consider the work done in \parencite{IJAIN54|to_array:0}, where the authors use K-means-based feature reduction and Naive Bayes classification pipeline. The approach operates on the 561-dimensional handcrafted features provided with the UCI HAR dataset \parencite{anguita2013public}. The study reports a baseline accuracy of $76.77\%$ using all 561 features and a maximum of 83.35\% after feature reduction (achieved at $k=70$).\\

However, this study provides limited implementation details and does not make its code publicly available, which prevents full reproducibility and limits rigorous comparison. Furthermore, it does not explicitly capture higher-order temporal interactions between sensor channels and lacks a fully specified and reproducible experimental pipeline. Moreover, the method based on K-means-based feature reduction followed by classification is very demanding from the algorithmic viewpoint for large values of $k$. 

\subsection{Proposed Strategy}

In this paper, we use the signature transform as a mathematically grounded tool for extracting structured features from temporal data streams. Coupled with a machine learning model, the proposed approach achieves an accuracy of $85.8\%$. The study follows a registered, pre-specified, and reproducible protocol, publicly available on the Open Science Framework (OSF)\footnote{\url{https://osf.io/9k4rc}}, ensuring transparency and enabling independent verification \parencite{jarrar2026protocol}. \\

The novelty of this study lies in a controlled evaluation of when signature-based representations improve HAR. We systematically examine the effects of embedding strategy, truncation depth, classifier choice, and sensor configuration under a unified, leakage-aware, and reproducible protocol. Our results show that the benefit of signatures is conditional: they improve performance with Random Forest, but not with MLP, demonstrating that their effectiveness depends on the interaction between representation and classifier. To the best of our knowledge, no prior study has systematically evaluated these factors jointly for signature-based feature learning on the UCI HAR dataset. This absence limits the ability to rigorously assess whether structured representations such as signatures provide a genuine advantage over conventional feature-based or model-based approaches on the UCI HAR dataset \parencite{anguita2013public}. \\

The main contributions of this study are: (i) a controlled comparison of raw, time-augmented, and lead--lag representations; (ii) a systematic analysis of the interaction between signature depth and classifier choice; (iii) a leakage-aware, subject-disjoint evaluation; and (iv) a fully reproducible framework supported by a pre-registered protocol, public code, model documentation, and a Docker environment.

\subsection{Plan of the Paper}

Our paper is organized as follows: \\

Section~\ref{signa} introduces the mathematical framework used throughout
the paper. We first define temporal data streams, their embeddings and
augmentations, and then introduce the signature of a continuous path,
its computation and main properties, and its extension to temporal
data streams.\\

Section~\ref{har} presents our signature-based feature approach for human
activity recognition. We first describe the structure of the HAR dataset
and then introduce the main strategy based on signature feature extraction,
different embeddings, truncation depths, and learning models.
Subsection~\ref{main} presents the experimental results across the different
models, representations, and configurations, followed by a class-level
evaluation of the best-performing model.\\

Section~\ref{benchmark} compares our approach with the K-means-based feature
reduction and Naive Bayes classification method proposed by
\textcite{IJAIN54|to_array:0}. Since the original implementation is not fully
specified, we reproduce the method in Appendix~\ref{ismi} using the
information provided in the original study.\\

Section~\ref{discussion} discusses the main findings, with particular
attention to the effects of representation choice, truncation depth, model
choice, and sensor configuration. It also discusses the strengths and
limitations of the study.\\

Finally, Section~\ref{conclusion} summarizes the main conclusions and
outlines perspectives for future work. Details concerning the dataset and the different data representations are
provided in Appendix~\ref{imple}. Model implementation details, the training
protocol, evaluation metrics, and the reproducibility framework are provided
in Appendix~\ref{modelimple}.

\section{Mathematical Framework: Temporal Data Streams and Signatures}
\label{signa}
This section introduces the mathematical framework underlying our approach.
We first define temporal data streams, their embeddings and augmentations,
and then introduce the signature of a continuous path and its main
properties. Finally, we extend the notion of signature to temporal data
streams. \\
\subsection{Temporal Data Streams}

Let $[a,b]\subset\mathbb{R}$ be a given interval and let $N\in\mathbb{N}$.
We consider the discrete time scale
\[
\mathbb{T}=\{t_0,t_1,\ldots,t_N\},
\qquad
a=t_0<t_1<\cdots<t_N=b.
\]

For $d\in\mathbb{N}$, a $d$-dimensional temporal data stream over
$\mathbb{T}$ is a map
\[
\mathbf{x}:\mathbb{T}\rightarrow\mathbb{R}^d.
\]
Equivalently, $\mathbf{x}$ is determined by $d$ discrete component functions
$x_j:\mathbb{T}\rightarrow\mathbb{R}$, $j=1,\ldots,d$, such that
\begin{equation}
\mathbf{x}(t)=\bigl(x_1(t),\ldots,x_d(t)\bigr),
\qquad t\in\mathbb{T}.
\end{equation}

We denote by $\mathscr{F}(\mathbb{T},\mathbb{R}^d)$ the set of all
$d$-dimensional temporal data streams over $\mathbb{T}$.

\subsection{Embedding of Temporal Data Streams}
\label{embed}

Each $d$-dimensional temporal stream can be associated with a $d$-dimensional continuous stream $\bX:[a,b]\rightarrow\R^d$. We call this process an {\bf embedding}. The embedding process is not unique. We will discuss two of them below: the \textbf{classical} and \textbf{lead-lag} embeddings.  \\

The {\bf standard embedding} denoted by $\mbox{Emb}_1$ is obtained by linear interpolation on $\bx$. Precisely, let us denote by $P^1_{\T} ([a,b], \R^d )$ the set of continuous piecewise linear (affine) functions with breakpoints in $\T$, which are linear on each interval $[t_i,t_{i+1}]$ and continuous on $[a,b]$.\\

Let $\mathbf{x}:\mathbb{T}\rightarrow\mathbb{R}^d$ be a given
$d$-temporal stream. The linear interpolation of $\mathbf{x}$,
denoted by $\mathbf{X}:[a,b]\rightarrow\mathbb{R}^d$, is defined,
for all $t\in[t_i,t_{i+1}]$, by
\begin{equation}
\label{linear}
\mathbf{X}(t)
=
\mathbf{x}(t_i)
+
\frac{t-t_i}{t_{i+1}-t_i}
\left(
\mathbf{x}(t_{i+1})-\mathbf{x}(t_i)
\right),
\qquad i=0,\ldots,N-1.
\end{equation}

We have $\mbox{Emb}_1 (\bx ) =\bX$.\\

We denote by $\mathscr{F} ([a,b],\R^d )$ the set of continuous $d$-dimensional temporal streams on $[a,b]$.\\

The {\bf lead-lag embedding} denoted by $\mbox{Emb}_2$ is defined as follows:\\

Let $\bx$ be a $d$-dimensional temporal stream. We denote by $\T^+$ (resp. $\T^-$) the discrete time scale $\T^+ =\T \setminus \{ b\}$ and $\T^- = \T \setminus \{ a \}$. We denote by $\sigma :\T^+ \rightarrow \T^-$ the shift operator defined for all $t_i\in\T^+$ by 
$$\sigma (t_i )=t_{i+1} .$$
We denote by $\bx^{\sigma}$ the $d$-dimensional temporal stream defined for all $t\in \T^+$ by 
$$\bx^{\sigma} (t) =\bx (\sigma (t)) .$$
The idea of the lead-lag embedding is to include more information than the present time to study the temporal data stream. The extended forward temporal stream associated with $\bx$ is the $2d$-dimensional temporal stream defined for all $t\in\T^+$ by 
$$(\bx ,\bx^{\sigma} ) .$$
The lead-lag embedding of $\bx$, denoted by $\mbox{Emb}_2(\bx)$, is then defined as 
\begin{equation}
\mbox{Emb}_2 (\bx )=\mbox{Emb}_1 (\bx ,\bx^{\sigma} ) .
\end{equation}

\subsection{Augmentation of Temporal Data Streams}

Let $\bX$ be a function in $\mathscr{F} ([a,b],\R^d )$. The time-augmented version of $\bX$, denoted by $\aX$, is the map from $[a,b]$ to $\R^{d+1}$ defined for all $t\in[a,b]$ by 
$$\mathbb{X}(t)=(t,\bX(t)).$$  
Augmentation can already be defined on the discrete level. Let $\bx$ be a $d$-dimensional temporal stream. We denote by ${x}$ the $(d+1)$-dimensional temporal stream defined for all $t\in \T$ by 
$${x} (t) = (t,\bx (t)).$$
We have 
$$\mbox{Emb}_1 ({x} ) (t) =(t,\mbox{Emb}_1 (\bx ) (t) ) ,$$
which coincides with the augmentation of $\mbox{Emb}_1 (\bx )$. As a consequence, augmentation can be done indifferently at the discrete or continuous level.\\

As we will see, the augmented version of the continuous temporal stream has interesting properties with respect to the signature that we introduce in the next section. \\

\subsection{Signature of a Continuous Path}

In this subsection, we recall some classical results on the signature of a continuous path. We refer to \parencite{lyons1998differential,Chevyrev_2025} for a review. As we are considering paths obtained by linear interpolation of $d$-dimensional data streams, we adopt a computational presentation, meaning that we focus on the explicit definition of the signature when the path is a continuous piecewise linear function. \\

Let $\bX:[a,b]\rightarrow \R^d$ be a family of $d$ functions defined on $[a,b]$. For each multi-index $\bi = (i_1 ,\dots ,i_k) \in \{ 1,\dots ,d\}^k$, we denote by $S_{a,b}^{\bi} (\bX )$ the quantity
\begin{equation}
S_{a,b}^{\bi} (\bX )=\int_{a}^b \int_a^{s_{k-1}} \dots \int_a^{s_1} dx_{i_1 ,s_1} \dots dx_{i_k,s_k}     
\end{equation}
called an iterated integral. We explicitly define this quantity in the following when $\bX$ is a continuous piecewise linear function.\\

For each $k\in \N$, we denote by $S_{a,b}^{(k)} \in \R^{d^k}$ the collection of coefficients $S_{a,b}^{\bi}$, $\bi \in\{ 1,\dots ,d\}^k$. This quantity is a {\bf tensor}: Let $e_1 = (1,0,\dots ,0), \dots ,e_d = (0,\dots ,1 )$ be the canonical basis in $\R^d$ and $\otimes$ the classical tensor product. We define by $S(\bX )$ the formal series 
\begin{equation}
    S_{a,b} (\bX )=1+\di\sum_{k\geq 1} \di\sum_{\bi =(i_1 ,\dots ,i_k ) \in \{ 1,\dots ,d\}^k} S_{a,b}^{\bi} (\bX) e_{i_1}\otimes \dots e_{i_k} ,
\end{equation}
which belongs to the tensor algebra $T((\R^d))$ of $\R^d$ and is called the {\bf signature} of $\bX$ over $[a,b]$.\\

In practice, we have to truncate the signature to a given {\bf depth} $m$, i.e., we consider 
\begin{equation}
    S_{a,b,m} (\bX )=1+\di\sum_{k=1}^{m}
    \di\sum_{\bi =(i_1 ,\dots ,i_k ) \in \{ 1,\dots ,d\}^k}
    S_{a,b}^{\bi} (\bX) e_{i_1}\otimes \dots \otimes e_{i_k} ,
\end{equation}
which can be identified with a vector of dimension
\begin{equation}
D(d,m) = \sum_{k=1}^{m} d^k.
\label{eq:dimension}
\end{equation}

\subsection{Explicit Computation of the Signature for Piecewise Linear Functions}

The explicit computation of the signature is easily done in the case of continuous piecewise linear functions. Indeed, in this case the integral $\di\int_a^t dX_s$ is defined for a real-valued affine function $X:[a,b]\rightarrow \R$ defined by 
\begin{equation}
\label{simple}
X(t)=\alpha t +\beta
\end{equation}
as
$$\di\int_a^t dX_s =\alpha (t-a).$$

If $\bX =\{ X_i \}_{i=1,\dots ,d}$ is a finite family of $d$ piecewise linear functions as in \eqref{simple} with coefficients $(\alpha_i ,\beta_i )_{i=1,\dots ,d}$, then we have 
\begin{equation}
    S_{a,b}^{\bi} (\bX ) = \di\frac{(b-a)^k}{k!} \alpha_{i_1} \dots \alpha_{i_k} .
\end{equation}
With the previous result, we are only able to compute the signature components $S_{u,v}^{\bi} (\bX )$ for a general continuous piecewise linear function as in \eqref{linear}, on each interval $[t_i ,t_{i+1}]$, $i=1,\dots ,N-1$. Fortunately, there exists a formula that allows us to compute the signature of a {\bf concatenated} function:\\

Let $X:[a,b]\rightarrow\R^d$ and $Y:[b,c]\rightarrow\R^d$ be two continuous functions. We denote by $X\star Y$ the concatenation of $X$ and $Y$, defined on $[a,c]$ by $(X\star Y) (t)=X(t)$ for $t\in [a,b]$ and $(X\star Y) (t)=X(b) + (Y(t)-Y(b))$ for $t\in [b,c]$.\\

The computation of $S_{a,c} (X\star Y)$ can then be obtained using {\bf Chen's formula}: 
\begin{equation}
    S_{a,c} (X\star Y ) = S_{a,b} (X)\otimes S_{b,c} (Y) ,
\end{equation}
and 
\begin{equation}
    S_{a,c}^{(k)} (X\star Y ) = \di\sum_{j=0}^k S_{a,b}^{(j)} (X)\otimes S_{b,c}^{(k-j)} (Y) ,
\end{equation}

\subsection{Properties of the Signature}

Let $\bX$ be a continuous function from $[a,b]$ to $\R^d$. Under some mild conditions, always satisfied by augmented continuous $d$-flux, the signature completely characterizes $\bX$, i.e., we have a one-to-one correspondence between $\bX$ and $S_{a,b}(\bX)$. \\

Moreover, we have the following two interesting properties when dealing with particular signals :\\

- {\bf Translation invariance}: Let $\gamma \in \R^d$, then $S_{a,b} (\bX +\gamma ) =S_{a,b} (\bX )$.\\

- {\bf Scaling property}: Let $\lambda\in \R$, for each $k\geq 1$, $S_{a,b}^{(k)} (\lambda \bX )= \lambda^k S_{a,b}^{(k)} (\bX )$.\\

These properties allow us to simplify computations in the case of periodic signals and signals possessing scaling properties.

\subsection{Signature of a $d$-Temporal Data Stream}

In this subsection, we extend the notion of signature to discrete $d$-dimensional temporal data streams. \\

Let $\bx$ be a $d$-dimensional temporal data stream and $\mbox{Emb}$ be a given embedding, i.e., a map from $\mathscr{F} (\T ,\R^d )$ into $\mathscr{F} ([a,b] ,\R^e)$ where $e\geq d$. For each $\bx \in \mathscr{F} (\T ,\R^d )$, we denote by 
$\mbox{Emb} (\bx )$ the associated embedded function. As embedding is a one-to-one process, we can define the signature of $\bx$ with respect to a given embedding $\mbox{Emb}$ as 
\begin{equation}
    S_{a,b}^{Emb} (\bx ) = S_{a,b} (\mbox{Emb} (\bx) ) .
\end{equation}
The choice of the embedding process is important and will be discussed.

\section{Signature-Based Features for Human Activity Recognition}
\label{har}
This section presents our signature-based approach for human activity
recognition. We first describe the HAR dataset and the proposed feature
extraction strategy. We then evaluate the different representations and
learning models and analyse the performance of the best configuration. \\
\subsection{Structure of the HAR Dataset}

The HAR dataset consists of labelled temporal data streams obtained from
inertial sensor measurements. We denote by $\D$ the collection of $p$
temporal data streams
\[
\bx_i:\T\rightarrow\R^d,
\qquad i=1,\ldots,p,
\]
where $p$ denotes the number of observations (windows), $N=127$, and
$d=6$. The six channels correspond to three-axis body acceleration and
three-axis gyroscope measurements. The dataset was collected from 30
participants. \\

We consider the set $\A$ of six activity classes: walking, walking
upstairs, walking downstairs, sitting, standing, and laying. For each
data stream $\bx_i\in\D$, we denote by $a_i\in\A$ its corresponding
activity label. The activities can be divided into two categories:
dynamic activities, consisting of walking, walking upstairs, and walking
downstairs, and static activities, consisting of sitting, standing, and
laying. The dataset can therefore be represented as a family of labelled
observations
\[
(\bx_i,a_i)\in\D\times\A,
\qquad i=1,\ldots,p.
\]

Following the predefined partition of the HAR dataset, the observations
are divided into training and test sets, denoted by
$(\D_{\mathrm{train}},\A_{\mathrm{train}})$ and
$(\D_{\mathrm{test}},\A_{\mathrm{test}})$, respectively. The HAR
classification problem consists of predicting the activity label
$a(\bx)$ associated with a given temporal data stream $\bx$. \\

In addition to the six-channel representation, we consider a reduced
dataset $\D_r$ containing only the three-axis body acceleration signals.
Each observation in this reduced representation therefore has $N=127$
and $d=3$. This allows us to evaluate the contribution of the gyroscope
signals by comparing accelerometer-only and full inertial representations.

\subsection{Main Strategy}
\label{strategy}

\begin{figure}[!ht]
\centering
\includegraphics[width=0.7\linewidth]{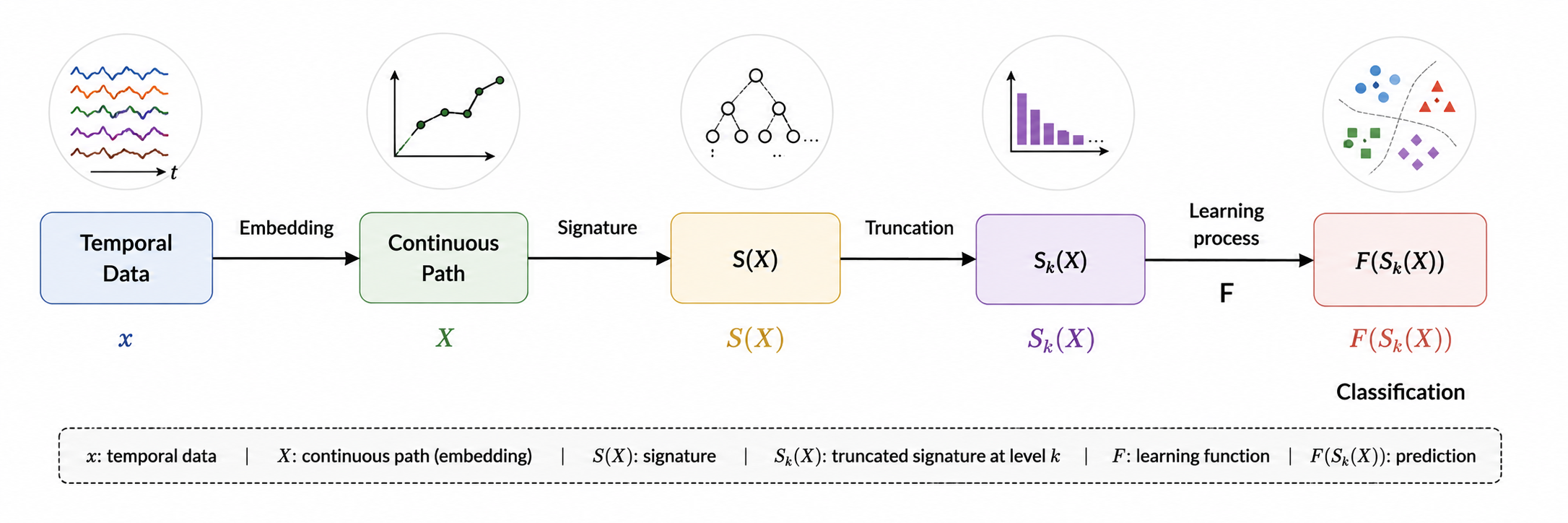}
\caption{Overview of the proposed signature-based feature learning and
classification pipeline. A temporal data stream $\mathbf{x}$ is mapped
to a continuous path $\mathbf{X}$ through an embedding procedure. The
path signature $S(\mathbf{X})$ is then computed and truncated at depth
$k$ to obtain $S_k(\mathbf{X})$. Finally, a learning function $F$ maps
the truncated signature representation to the predicted activity class.}
\label{fig:methodology_pipeline}
\end{figure}

Let $\mathbf{x}$ be a $d$-temporal data stream and $\mbox{Emb}$ a given
embedding. For a given truncation depth $k$, we denote by
$S_{k,(a,b)}^{Emb}(\mathbf{x})$ the truncated signature of $\mathbf{x}$
up to depth $k$. The proposed strategy, illustrated in
Figure~\ref{fig:methodology_pipeline}, consists of embedding the temporal
data stream into a continuous path, computing its truncated signature,
and using the resulting signature features as input to a classifier. \\

The proposed method is mainly determined by two elements: the choice of
embedding and the truncation depth of the signature. We consider the
time-augmented and lead--lag embeddings defined in
Section~\ref{embed}, with truncation depths up to $k=6$.
Different embedding choices can lead to substantial differences in
classification performance even when the truncation depth is fixed
\parencite{FERMANIAN2021107148}. Time augmentation introduces a strictly
increasing time channel that preserves ordering information, whereas the
lead--lag embedding represents delayed and current values jointly,
thereby exposing temporal interactions that may otherwise remain hidden
\parencite{FERMANIAN2021107148}. These properties are particularly
relevant for inertial HAR, where accelerometer and gyroscope signals may
exhibit coordinated temporal dynamics \parencite{hammerla2016,bulling2014tutorial}. \\

Unlike heuristic feature engineering or implicit representation learning,
signatures provide a structured representation of temporal interactions,
with higher-order terms capturing increasingly complex dependencies
across time \parencite{lyons1998differential}. For multivariate signals,
the complexity of this representation can be explicitly controlled
through the truncation depth \parencite{Chevyrev_2025}. In particular,
suitable embeddings can make informative temporal interactions accessible
at relatively low truncation depths
\parencite{FERMANIAN2021107148,Chevyrev_2025}. \\

We evaluate the resulting representations using two classifiers:
a multilayer perceptron (MLP) and a Random Forest (RF). The MLP captures
nonlinear relationships through parametric function approximation
\parencite{Taud2018}, whereas the RF provides a nonparametric ensemble
model capable of capturing complex feature interactions
\parencite{breiman2001random}. We evaluate the effects of embedding
choice and truncation depth under a unified experimental protocol,
allowing comparison across signature representations, raw representations,
and model classes. Implementation details of the experimental protocol
are provided in Appendix~\ref{imple}.

\subsection{Performance Across Models, Representations, and Configurations}
\label{main}

The performance of all evaluated models is assessed under a unified, leakage-aware experimental framework and is summarised in Tables~1--3. Metrics include validation macro F1, test accuracy, test macro F1 score, corresponding 95\% confidence intervals, and classification counts. All models are evaluated using identical data splits and preprocessing procedures. \\

\subsubsection{MLP Models with Signature-Based Representations}

The performance of multilayer perceptron (MLP) models trained on signature-based representations derived from time augmentation and lead--lag transformations is summarised in Table~\ref{tab:mlp_signature}. Experiments are conducted using both six-channel (full inertial) and three-channel (accelerometer-only) inputs, with truncation depths from 2 to 4 and activation functions including ReLU and Tanh. \\

\begin{table*}[t]
\caption{Performance of MLP models with signature-based representations across configurations. All reported metrics include validation performance, test performance with 95\% confidence intervals, and classification counts.}
\label{tab:mlp_signature}
\centering
\resizebox{\textwidth}{!}{
\begin{tabular}{llllllllllllllll}
\toprule
Rep. & Ch. & Depth & Dim. & Epoch & Val F1 & Acc & Acc CI & F1 & F1 CI & Correct & Incorrect & Corr (\%) & Incorr (\%) & Act. \\
\midrule

Time Aug. & 6 & 2 & 56 & 26 & 0.622 & 0.599 & [0.582--0.616] & 0.607 & [0.591--0.623] & 1764 & 1183 & 59.857 & 40.143 & ReLU \\
Time Aug. & 6 & 3 & 399 & 12 & 0.648 & 0.622 & [0.606--0.639] & 0.635 & [0.621--0.649] & 1834 & 1113 & 62.233 & 37.767 & ReLU \\
Time Aug. & 6 & 4 & 2800 & 7 & 0.645 & 0.628 & [0.611--0.645] & 0.631 & [0.617--0.645] & 1852 & 1095 & 62.844 & 37.156 & ReLU \\

Lead--Lag & 6 & 2 & 210 & 19 & 0.621 & 0.616 & [0.598--0.633] & 0.629 & [0.613--0.644] & 1815 & 1132 & 61.588 & 38.412 & ReLU \\
Lead--Lag & 6 & 3 & 2954 & 5 & 0.625 & 0.652 & [0.635--0.669] & 0.637 & [0.624--0.651] & 1920 & 1027 & 65.151 & 34.849 & ReLU \\
Lead--Lag & 6 & 4 & 41370 & 18 & 0.650 & 0.639 & [0.622--0.656] & 0.628 & [0.615--0.641] & 1883 & 1064 & 63.895 & 36.105 & ReLU \\

Time Aug. & 3 & 2 & 20 & 10 & 0.534 & 0.520 & [0.502--0.537] & 0.529 & [0.512--0.546] & 1533 & 1414 & 52.019 & 47.981 & ReLU \\
Time Aug. & 3 & 3 & 84 & 16 & 0.661 & 0.678 & [0.661--0.694] & 0.689 & [0.672--0.704] & 1997 & 950 & 67.764 & 32.236 & ReLU \\
Time Aug. & 3 & 4 & 340 & 42 & 0.743 & 0.734 & [0.718--0.750] & 0.741 & [0.725--0.755] & 2164 & 783 & 73.431 & 26.569 & ReLU \\

Lead--Lag & 3 & 2 & 72 & 24 & 0.603 & 0.635 & [0.618--0.653] & 0.628 & [0.612--0.643] & 1872 & 1075 & 63.522 & 36.478 & ReLU \\
Lead--Lag & 3 & 3 & 584 & 14 & 0.699 & 0.721 & [0.705--0.738] & 0.729 & [0.714--0.744] & 2126 & 821 & 72.141 & 27.859 & ReLU \\
Lead--Lag & 3 & 4 & 4680 & 26 & 0.690 & 0.688 & [0.671--0.705] & 0.668 & [0.654--0.682] & 2028 & 919 & 68.816 & 31.184 & ReLU \\

Time Aug. & 3 & 2 & 20 & 10 & 0.511 & 0.490 & [0.472--0.508] & 0.482 & [0.467--0.498] & 1443 & 1504 & 48.965 & 51.035 & Tanh \\
Time Aug. & 3 & 3 & 84 & 19 & 0.621 & 0.603 & [0.586--0.620] & 0.616 & [0.600--0.631] & 1777 & 1170 & 60.299 & 39.701 & Tanh \\
Time Aug. & 3 & 4 & 340 & 3 & 0.627 & 0.585 & [0.569--0.603] & 0.573 & [0.558--0.588] & 1723 & 1224 & 58.466 & 41.534 & Tanh \\

Lead--Lag & 3 & 2 & 72 & 6 & 0.593 & 0.587 & [0.570--0.604] & 0.576 & [0.561--0.591] & 1731 & 1216 & 58.738 & 41.262 & Tanh \\
Lead--Lag & 3 & 3 & 584 & 26 & 0.656 & 0.650 & [0.634--0.668] & 0.663 & [0.648--0.678] & 1916 & 1031 & 65.015 & 34.985 & Tanh \\
Lead--Lag & 3 & 4 & 4680 & 33 & 0.650 & 0.651 & [0.635--0.668] & 0.653 & [0.638--0.667] & 1919 & 1028 & 65.117 & 34.883 & Tanh \\

\bottomrule
\end{tabular}
}
\end{table*}

For six-channel inputs with ReLU activation function, MLP performance remains within a relatively narrow range across all evaluated configurations. Time-augmented representations yield test accuracies from 0.599 to 0.628 and macro F1 scores from 0.607 to 0.631. Lead--lag representations achieve the highest performance within this group, with test accuracies ranging from 0.616 to 0.652 and macro F1 scores from 0.628 to 0.637. The best-performing six-channel configuration corresponds to the lead--lag representation at depth 3, achieving a test accuracy of 0.652 and a macro F1 score of 0.637. \\

For three-channel inputs with ReLU activation function, performance increases with truncation depth up to depth 4 for time-augmented representations and up to depth 3 for lead--lag representations, followed by a decline at depth 4 for lead--lag. Time-augmented representations improve from a test accuracy of 0.520 at depth 2 to 0.734 at depth 4, with macro F1 scores increasing from 0.529 to 0.741. Lead--lag representations yield test accuracy values of 0.635, 0.721, and 0.688 at depths 2, 3, and 4, respectively, with corresponding macro F1 scores of 0.628, 0.729, and 0.668. The strongest three-channel configuration is obtained using time augmentation at depth 4, with 2164 correctly classified instances (73.431\%). Notably, this configuration achieves higher performance than all six-channel MLP signature-based settings. \\

For three-channel inputs with Tanh activation function, performance is consistently lower than with ReLU across both representation types. Time-augmented representations achieve test accuracy values between 0.490 and 0.603 and macro F1 scores between 0.482 and 0.616. Lead--lag representations yield test accuracy values between 0.587 and 0.651 and macro F1 scores between 0.576 and 0.663. The highest-performing configuration within this group is the lead--lag representation at depth 4, achieving a test accuracy of 0.651 and a macro F1 score of 0.653. \\

Across all MLP configurations, input dimensionality increases substantially with truncation depth, particularly for lead--lag representations. In the six-channel setting, dimensionality ranges from 56 to 2800 for time augmentation and from 210 to 41{,}370 for lead--lag. In the three-channel setting, dimensionality ranges from 20 to 340 for time augmentation and from 72 to 4680 for lead--lag. 

\subsubsection{Random Forest Models with Signature-Based Representations}

The performance of Random Forest (RF) models trained on signature-based representations using six-channel inputs is presented in Table~\ref{tab:rf_signature}. Both Gini and entropy splitting criteria are evaluated across time-augmented and lead--lag representations. \\

\begin{table*}[t]
\caption{Performance of Random Forest (RF) models with signature-based representations across configurations. Results are reported for both Gini and entropy splitting criteria, including validation performance, test performance with 95\% confidence intervals, and classification counts.}
\label{tab:rf_signature}
\centering
\resizebox{\textwidth}{!}{
\begin{tabular}{llllllllllllllll}
\toprule
Rep. & Ch. & Depth & Dim. & Epoch & Val F1 & Acc & Acc CI & F1 & F1 CI & Correct & Incorrect & Corr (\%) & Incorr (\%) & Criterion \\
\midrule

Time Aug. & 6 & 2 & 56 & 33 & 0.801 & 0.767 & [0.752--0.782] & 0.771 & [0.756--0.785] & 2260 & 687 & 76.688 & 23.312 & Gini \\
Time Aug. & 6 & 3 & 399 & 39 & 0.838 & 0.800 & [0.786--0.814] & 0.803 & [0.788--0.817] & 2358 & 589 & 80.014 & 19.986 & Gini \\
Time Aug. & 6 & 4 & 2800 & 49 & 0.865 & 0.828 & [0.814--0.841] & 0.830 & [0.816--0.844] & 2441 & 506 & 82.830 & 17.170 & Gini \\

Lead--Lag & 6 & 2 & 210 & 29 & 0.800 & 0.777 & [0.763--0.792] & 0.781 & [0.766--0.795] & 2291 & 656 & 77.740 & 22.260 & Gini \\
Lead--Lag & 6 & 3 & 2954 & 21 & 0.827 & 0.804 & [0.790--0.818] & 0.807 & [0.792--0.821] & 2370 & 577 & 80.421 & 19.579 & Gini \\
Lead--Lag & 6 & 4 & 41370 & 28 & 0.855 & 0.831 & [0.817--0.844] & 0.833 & [0.819--0.846] & 2449 & 498 & 83.101 & 16.899 & Gini \\

Time Aug. & 6 & 2 & 56 & 44 & 0.826 & 0.775 & [0.761--0.789] & 0.778 & [0.763--0.792] & 2284 & 663 & 77.503 & 22.497 & Entropy \\
Time Aug. & 6 & 3 & 399 & 41 & 0.852 & 0.796 & [0.782--0.810] & 0.798 & [0.783--0.812] & 2347 & 600 & 79.640 & 20.360 & Entropy \\
Time Aug. & 6 & 4 & 2800 & 50 & 0.871 & 0.838 & [0.825--0.852] & 0.840 & [0.827--0.853] & 2471 & 476 & 83.848 & 16.152 & Entropy \\
Time Aug. & 6 & 5 & 19607 & 45 & 0.870 & 0.845 & [0.832--0.858] & 0.847 & [0.833--0.859] & 2489 & 458 & 84.459 & 15.541 & Entropy \\
Time Aug. & 6 & 6 & 137256 & 49 & 0.874 & 0.858 & [0.845--0.871] & 0.859 & [0.845--0.870] & 2529 & 418 & 85.816 & 14.184 & Entropy \\

Lead--Lag & 6 & 2 & 210 & 9 & 0.792 & 0.766 & [0.750--0.782] & 0.771 & [0.755--0.785] & 2256 & 691 & 76.552 & 23.448 & Entropy \\
Lead--Lag & 6 & 3 & 2954 & 32 & 0.830 & 0.822 & [0.808--0.835] & 0.825 & [0.810--0.838] & 2421 & 526 & 82.151 & 17.849 & Entropy \\
Lead--Lag & 6 & 4 & 41370 & 49 & 0.861 & 0.839 & [0.827--0.852] & 0.841 & [0.828--0.854] & 2474 & 473 & 83.950 & 16.050 & Entropy \\

\bottomrule
\end{tabular}
}
\end{table*}

Under the Gini criterion, performance improves consistently with increasing truncation depth for both representation types. Time-augmented representations yield test accuracies from 0.767 at depth 2 to 0.828 at depth 4, with macro F1 scores increasing from 0.771 to 0.830. Lead--lag representations achieve test accuracies from 0.777 to 0.831 and macro F1 scores from 0.781 to 0.833 across the same depths. The highest performance under the Gini criterion is obtained using the lead--lag representation at depth 4, achieving a test accuracy of 0.831 and a macro F1 score of 0.833. \\

Under the entropy criterion, time-augmented representations are evaluated across a wider range of truncation depths (2--6), with performance increasing monotonically with depth. Test accuracy improves from 0.775 at depth 2 to 0.858 at depth 6, while macro F1 increases from 0.778 to 0.859. The corresponding number of correctly classified instances rises from 2284 to 2529. The highest-performing configuration across all RF experiments is obtained using time-augmented representations at depth 6, achieving a test accuracy of 0.858 and a macro F1 score of 0.859. \\

Lead--lag representations under the entropy criterion yield test accuracies of 0.766, 0.822, and 0.839 at depths 2, 3, and 4, respectively, with corresponding macro F1 scores of 0.771, 0.825, and 0.841. Performance increases with depth up to depth 4 but remains below that of the best time-augmented configurations. \\

Across both splitting criteria, input dimensionality increases substantially with truncation depth, particularly for time-augmented representations evaluated up to depth 6. Dimensionality ranges from 56 to 137{,}256 for time augmentation and from 210 to 41{,}370 for lead--lag representations. Across all RF configurations, the number of correctly classified instances ranges from 2256 to 2529. 

\subsubsection{Baseline Models with Raw Flattened Signals}
\label{flat}

Let $\T$ be a given discrete time-scale on $[a,b]$ defined by $\T =\{ t_0 ,\dots ,t_N \}$ with $N\geq 2$. Let $p\in \N$, we consider a dataset $\D$ consisting of a family of $p$ data streams $\bx_i :\T \rightarrow \R^d$ with $d\in \N$. For each $i\in \{ 1,\dots ,p\}$ and $j\in \{ 1,\dots ,d \}$, we denote by $\bx_{i,j} :\T \rightarrow \R$ the components of $\bx_i$ and by $\mathbf{v}_{i,j} \in \R^{N+1}$ the vector defined by
$$\mathbf{v}_{i,j}  =(\bx_{i,j} (t_0) ,\dots ,\bx_{i,j} (t_N) ) .$$
The {\bf flattening} of the dataset $\D$ denoted by $\D_{flat}$ is the family of $p$ vectors $\mathbf{v}_i\in \R^{d(N+1)}$, $i\in \{1,\dots,p\}$ defined by
$$
\mathbf{v}_i=(\mathbf{v}_{i,1},\dots,\mathbf{v}_{i,d}).
$$
Note that, in this representation, the temporal structure of the data stream is no longer explicitly encoded.\\

The baseline performance of models trained on raw flattened sensor signals is reported in Table~\ref{tab:baseline}.\\

\begin{table*}[t]
\caption{Baseline performance of MLP and Random Forest (RF) models trained on raw flattened sensor signals without explicit temporal encoding. Metrics include validation performance, test performance with 95\% confidence intervals, and classification counts.}
\label{tab:baseline}
\centering
\resizebox{\textwidth}{!}{
\begin{tabular}{llllllllllllllll}
\toprule
Model & Rep. & Ch. & Dim. & Epoch & Val F1 & Acc & Acc CI & F1 & F1 CI & Correct & Incorrect & Corr (\%) & Incorr (\%) & Act./Criterion \\
\midrule

MLP & Raw & 6 & 768 & 33 & 0.749 & 0.794 & [0.780--0.808] & 0.799 & [0.786--0.813] & 2341 & 606 & 79.437 & 20.563 & ReLU \\
MLP & Raw & 3 & 384 & 43 & 0.679 & 0.769 & [0.753--0.784] & 0.772 & [0.758--0.786] & 2266 & 681 & 76.892 & 23.108 & ReLU \\
MLP & Raw & 3 & 384 & 23 & 0.539 & 0.626 & [0.609--0.644] & 0.639 & [0.624--0.654] & 1846 & 1101 & 62.640 & 37.360 & Tanh \\

RF & Raw & 6 & 768 & 46 & 0.772 & 0.816 & [0.802--0.830] & 0.817 & [0.803--0.830] & 2404 & 543 & 81.574 & 18.426 & Gini \\
RF & Raw & 6 & 768 & 41 & 0.769 & 0.812 & [0.797--0.826] & 0.814 & [0.801--0.827] & 2393 & 554 & 81.201 & 18.799 & Entropy \\

\bottomrule
\end{tabular}
}
\end{table*}

For MLP models with ReLU activation function, six-channel inputs achieve a test accuracy of 0.794 and a macro F1 score of 0.799, corresponding to 2341 correctly classified instances. Using three-channel inputs, performance decreases to a test accuracy of 0.769 and a macro F1 score of 0.772. For three-channel inputs with Tanh activation function, performance is substantially lower, with a test accuracy of 0.626 and a macro F1 score of 0.639.\\

For Random Forest models, six-channel inputs achieve test accuracies of 0.816 under the Gini criterion and 0.812 under the entropy criterion, with corresponding macro F1 scores of 0.817 and 0.814. The highest-performing baseline configuration is obtained using Random Forest with the Gini criterion, achieving 2404 correctly classified instances (81.574\%).\\

Across all baseline configurations, the proportion of correctly classified instances ranges from 62.640\% to 81.574\%. 

\subsubsection{Comparative Analysis Across Configurations}

Comparative results across activation functions, sensor configurations, representation types, and model classes are summarised in Tables~\ref{tab:mlp_signature}--\ref{tab:baseline}. \\

Across model classes, Random Forest consistently achieves higher performance than MLP across all representation types and configurations. The strongest overall performance is obtained by Random Forest models trained on time-augmented representations, with accuracy increasing monotonically with truncation depth under the entropy criterion and reaching a maximum of 0.858. \\

For MLP models, performance depends more strongly on representation and activation function. ReLU activation function consistently outperforms Tanh across all configurations. In the three-channel setting, time-augmented representations achieve the highest MLP performance, reaching a test accuracy of 0.734 at depth 4, while lead--lag representations peak at depth 3 before declining. \\

Comparisons across sensor configurations show that six-channel inputs generally yield higher performance for baseline models and Random Forest classifiers. However, in MLP models with signature-based representations, three-channel configurations can achieve higher peak performance than six-channel settings, particularly for time-augmented representations. \\

Across representation types, time-augmented representations consistently achieve the highest performance in Random Forest models, particularly at higher truncation depths. Lead--lag representations provide competitive performance at moderate depths but do not surpass the best time-augmented configurations. \\

Baseline comparisons show that raw representations achieve strong performance, particularly for Random Forest models, with a maximum test accuracy of 0.816. However, signature-based representations exceed baseline performance in the strongest configurations, particularly for Random Forest models using time-augmented representations at higher truncation depths. 

\subsection{Best Model Performance and Class-Level Evaluation}

The best-performing configuration corresponds to a Random Forest model trained on time-augmented representations with six-channel inputs at truncation depth $m=6$, using the entropy splitting criterion (Table~\ref{tab:rf_signature}). This model achieves a test accuracy of 0.858 (95\% CI: [0.845--0.871]) and a macro F1 score of 0.859 (95\% CI: [0.845--0.870]), with 2529 correctly classified instances (85.816\% of the test set). \\

The class-level performance in terms of precision, recall, F1-score, and support is summarised in Table~\ref{tab:class_performance}.

\begin{table}[t]
\caption{Class-level performance of the best-performing Random Forest model trained on time-augmented representations (depth $m=6$, entropy criterion). Metrics include precision, recall, F1-score, and support for each activity class.}
\label{tab:class_performance}
\centering
\begin{tabular}{lcccc}
\toprule
Class & Precision & Recall & F1-score & Support \\
\midrule

WALKING & 0.944 & 0.915 & 0.929 & 496 \\
WALKING\_UPSTAIRS & 0.871 & 0.934 & 0.902 & 471 \\
WALKING\_DOWNSTAIRS & 0.873 & 0.836 & 0.854 & 420 \\

SITTING & 0.779 & 0.768 & 0.773 & 491 \\
STANDING & 0.794 & 0.891 & 0.840 & 532 \\
LAYING & 0.906 & 0.806 & 0.853 & 537 \\

\midrule
Accuracy & -- & -- & 0.858 & 2947 \\
Macro Avg. & 0.861 & 0.858 & 0.859 & 2947 \\
Weighted Avg. & 0.861 & 0.858 & 0.858 & 2947 \\

\bottomrule
\end{tabular}
\end{table}

Among dynamical activities, \textit{WALKING} achieves the highest performance, with precision of 0.944, recall of 0.915, and F1-score of 0.929. \textit{WALKING\_UPSTAIRS} and \textit{WALKING\_DOWNSTAIRS} achieve F1-scores of 0.902 and 0.854, respectively. \\

For stationary activities, \textit{STANDING} achieves the highest recall (0.891) among these classes, with an F1-score of 0.840. \textit{LAYING} achieves high precision (0.906) with an F1-score of 0.853, while \textit{SITTING} yields the lowest performance overall, with an F1-score of 0.773. \\

Overall, classification performance is balanced across classes, with macro-averaged precision, recall, and F1-score of 0.861, 0.858, and 0.859, respectively. Weighted averages are 0.861 for precision, 0.858 for recall, and 0.858 for F1-score across 2947 test instances. \\

\begin{figure}[h]
\centering
\includegraphics[width=0.9\linewidth]{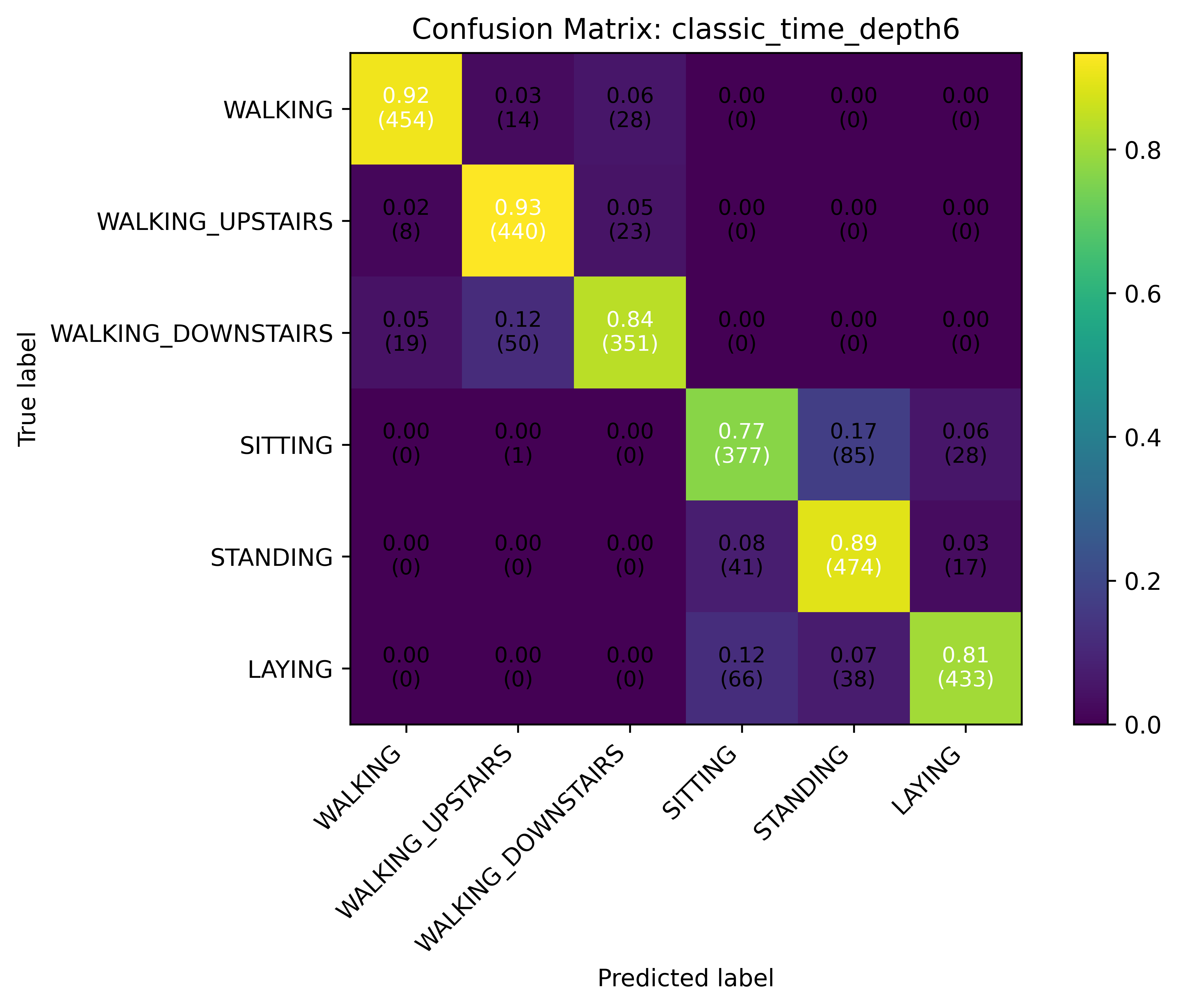}
\caption{Confusion matrix of the best-performing Random Forest model trained on time-augmented representations (depth $m=6$, entropy criterion). Values are shown as row-normalised proportions with corresponding raw counts in parentheses. Rows represent true labels and columns represent predicted labels.}
\label{fig:confusion_matrix_best}
\end{figure}

The confusion matrix for the best-performing model, reporting both raw counts and row-normalised values across all activity classes, is shown in Figure~\ref{fig:confusion_matrix_best}. 

\section{Comparison with Ismi et al. K-means-based Features}
\label{benchmark}

As explained in the introduction, the results of \parencite{IJAIN54|to_array:0} were not provided with sufficient information in order to check their method. We decided to reproduce their result by providing a detailed implementation in Appendix \ref{ismi}. Despite our effort, we were unable to obtain the performance indicated by the authors and obtain a weaker estimation. Precisely, Table~\ref{tab:reproduction} presents the results of the reproduced K-means-based feature reduction pipeline alongside those reported in the original study \parencite{IJAIN54|to_array:0}. \\

\begin{table}[t]
\caption{Reproduction results of the K-means-based feature reduction pipeline compared with the results reported in the original study \parencite{IJAIN54|to_array:0}. Performance is reported as correctly and incorrectly classified percentages across different cluster sizes ($k$).}
\label{tab:reproduction}
\centering
\begin{tabular}{llcc}
\toprule
Setting & Dataset & Correct (\%) & Incorrect (\%) \\
\midrule

\multirow{8}{*}{Reproduced} 
& Original HAR & 69.79 & 30.21 \\
& $k=10$ & 69.87 & 30.13 \\
& $k=20$ & 70.48 & 29.52 \\
& $k=30$ & 68.51 & 31.49 \\
& $k=40$ & 74.69 & 25.31 \\
& $k=50$ & 78.49 & 21.51 \\
& $k=60$ & 79.71 & 20.29 \\
& $k=70$ & 80.10 & 19.90 \\

\midrule

\multirow{8}{*}{Original Study \parencite{IJAIN54|to_array:0}}
& Original HAR & 76.77 & 23.23 \\
& $k=10$ & 69.40 & 30.60 \\
& $k=20$ & 76.81 & 23.19 \\
& $k=30$ & 79.56 & 20.44 \\
& $k=40$ & 79.33 & 20.67 \\
& $k=50$ & 81.38 & 18.62 \\
& $k=60$ & 79.84 & 20.16 \\
& $k=70$ & 83.35 & 16.65 \\

\bottomrule
\end{tabular}
\end{table}

In the reproduced implementation, the baseline configuration using all handcrafted features achieves a correctly classified percentage of 69.79\%. Applying K-means-based feature reduction improves performance progressively with increasing cluster size, reaching a maximum of 80.10\% at $k=70$. The corresponding incorrectly classified percentage decreases from 30.21\% to 19.90\%. \\

In the original study, the baseline performance is reported as 76.77\%. With feature reduction, performance increases to a maximum of 83.35\% at $k=70$. Across most values of $k$, the reproduced results are lower than those reported in the original study, with differences varying across configurations. \\

For comparison, the best-performing model proposed in this study achieves a correctly classified percentage of 85.816\% and an incorrectly classified percentage of 14.184\% on the test set. This exceeds both the reproduced implementation and the original study across all evaluated configurations. \\

\section{Discussion}
\label{discussion}
This section discusses the main findings of the study. We examine the
effects of representation, truncation depth, and model choice, compare our
results with a prior feature selection approach, and discuss the main
strengths and limitations of the proposed framework. \\
\subsection{Effect of Representation, Truncation Depth, and Model Choice}

This study provides a controlled evaluation of signature-based feature learning in human activity recognition (HAR), demonstrating that its effectiveness is real but conditional. Performance gains depend critically on representation design, truncation depth, and classifier choice. This leads to a central insight: representation should not be treated as a minor preprocessing step, but as a fundamental modelling decision that determines how temporal structure is made available to the classifier. This perspective aligns with prior work emphasising the importance of representation learning in time-series modelling \parencite{wang2019deep,FERMANIAN2021107148}.\\

Raw windows preserve original sensor measurements but do not explicitly encode temporal ordering or cross-channel interactions, leaving the classifier to infer these patterns implicitly \parencite{s16010115}. In contrast, signature-based representations shift part of this burden to the feature construction stage \parencite{Chevyrev_2025}. Time augmentation introduces a strictly increasing temporal dimension, preserving ordering \parencite{graham2013sparsearrayssignaturesonline}, while lead--lag embedding aligns delayed and current values, exposing interactions at lower orders \parencite{gyurkó2014extractinginformationsignaturefinancial}. As a result, the classifier operates on representations in which temporal structure is partially made explicit, rather than entirely learned. \\

The results show that explicit signature-based representations can improve HAR performance compared with raw-window baselines, although this improvement is classifier-dependent. For MLP models, raw flattened inputs outperformed all signature-based configurations, suggesting that the network effectively exploits the structured fixed-length input. In contrast, for Random Forest (RF), signature features led to clear improvements, with the best model achieving 0.858 accuracy and 0.859 macro F1, exceeding the strongest raw baseline (0.816 accuracy). This demonstrates that the primary advantage of signatures lies in providing structured, high-order summaries of temporal interactions that can be effectively exploited by nonparametric models \parencite{Chevyrev_2025}. More broadly, this suggests that signature-based representations are particularly effective when representation learning is decoupled from model training, rather than embedded within end-to-end optimisation \parencite{bengio2013representation}.\\

The comparison between time augmentation and lead--lag shows that both embeddings are beneficial, but their effectiveness differs. Lead--lag performs competitively at moderate depths, particularly in MLP settings, consistent with its ability to expose inter-channel interactions at lower orders \parencite{Chevyrev_2025}. However, time augmentation yields the highest overall performance, especially for RF, where accuracy improves monotonically with depth. This suggests that, for fixed-window HAR data, preserving temporal order with limited dimensional expansion provides a better balance between expressiveness and tractability than the more complex lead--lag embedding.\\

Truncation depth plays a significant role in performance \parencite{lyons2025}. For RF models, increasing depth consistently improves results, indicating that higher-order interactions remain informative. In contrast, MLP performance is less stable, with improvements plateauing or declining at higher depths. This reflects a trade-off between representational richness and learnability: deeper signatures increase expressiveness but also introduce high dimensionality that can hinder optimisation \parencite{10.1145/3371237}. Therefore, truncation depth should be treated as a key modelling hyperparameter rather than a technical detail \parencite{Chevyrev_2025}.\\

Across all configurations, RF consistently outperforms MLP under the leakage-aware evaluation protocol. This is likely due to its ability to handle high-dimensional, structured features without requiring complex optimisation \parencite{Iranzad2025}. Importantly, the use of subject-disjoint validation ensures that these gains reflect genuine generalisation rather than optimistic bias, strengthening the reliability of the findings.\\

Finally, accelerometer-only data can approximate full inertial performance to a considerable extent, particularly in MLP settings where lower dimensionality aids optimisation. However, the best overall results are obtained using six-channel inputs, indicating that gyroscope signals provide complementary information that improves classification.\\

In summary, the effectiveness of signature-based feature learning is not intrinsic, but emerges from the alignment between representation design, model capacity, and evaluation protocol. The highest results are achieved when structured temporal representations are combined with models capable of exploiting high-order interactions within a reproducible and leakage-aware framework. 

\subsection{Comparison with Prior Feature Selection Approaches}

The closest prior study on the UCI HAR dataset is the K-means-based feature reduction approach of \textcite{IJAIN54|to_array:0}, which clusters the original 561 handcrafted features and applies Naive Bayes classification. That study reported 76.77\% accuracy using all features and a maximum of 83.35\% at $k=70$, suggesting that reducing redundancy can improve performance. \\

Our reproduction confirms this general trend but yields lower performance, reaching 80.10\% at $k=70$. This discrepancy highlights an important limitation of the prior work: incomplete reporting of preprocessing, parameter settings, and validation procedures, which restricts exact reproducibility and introduces uncertainty regarding the reported results. \\

More importantly, this comparison clarifies the contribution of the present study. The K-means approach operates on fixed handcrafted features and focuses on redundancy reduction within a static representation. In contrast, our framework addresses a more fundamental problem: how temporal structure should be represented before classification. Signature-based representations explicitly encode ordering and inter-channel interactions, shifting the focus from feature selection to representation design \parencite{lyons1998differential,Chevyrev_2025}. \\

This distinction is reflected in performance. While the best reproduced and reported K-means results reach 80.10\% and 83.35\%, respectively, the best model in this study achieves 85.816\% on a held-out test set. This suggests that improvements in HAR are not driven solely by reducing feature redundancy, but by explicitly encoding temporal dynamics in a form that models can exploit. \\

Unlike the prior study, our framework is fully specified and publicly documented (Section~3.8), enabling transparent comparison and independent verification. Overall, these findings indicate that modelling temporal structure is more effective than feature reduction alone, and that performance gains are better understood as a consequence of improved representation rather than dimensionality optimisation. 

\subsection{Strengths and Limitations}

A key strength of this study is the use of a fully specified and reproducible experimental framework. All methodological decisions were defined in advance through a pre-registered protocol, available on the Open Science Framework (OSF)\footnote{\url{https://osf.io/9k4rc}}, and the complete pipeline—including preprocessing, feature construction, model training, and evaluation—is publicly available via a GitHub repository\footnote{\url{https://github.com/pwd529/har-signature-ML-reproducibility}}. Reproducibility is further supported through a Hugging Face model card\footnote{\url{https://huggingface.co/KamalJarrar/har-signature-rf-classic-time-depth6}} describing the best-performing model and a Docker image\footnote{\url{https://hub.docker.com/r/pwd529/har-signature-best-model}} enabling exact computational replication \parencite{10.5555/2600239.2600241,10.1145/3287560.3287596}. In addition, the use of a leakage-aware, subject-disjoint validation strategy ensures that reported performance reflects genuine generalisation rather than optimistic bias \parencite{10.1145/2382577.2382579}. \\

Another strength is the controlled evaluation design, which isolates the effect of representation, truncation depth, model choice, and sensor configuration under identical conditions. This allows for a fair and interpretable comparison across approaches \parencite{JMLR:v15:delgado14a,46555}. \\

However, several limitations should be noted. First, experiments are restricted to a single dataset, may limit generalisability to other domains or sensor configurations \parencite{Zhang2019ImproveMG}. Second, signature features can lead to high-dimensional representations, particularly at larger truncation depths, which may introduce computational challenges and affect model stability \parencite{Chevyrev_2025}. Finally, while two model families are considered, additional architectures (e.g., convolutional or recurrent models) were not explored, which may influence the relative performance of representations \parencite{hammerla2016}.

\section{Conclusion and Perspectives}
\label{conclusion}
This section summarises the main conclusions of the study and outlines
directions for future research on signature-based representations for
human activity recognition. \\
\subsection{Conclusion}

This study presented a fully reproducible and leakage-aware evaluation of signature-based feature learning for human activity recognition. By systematically comparing raw signals, time-augmented paths, and lead--lag embeddings across multiple truncation depths, model classes, and sensor configurations, the results demonstrate that the effectiveness of signature features is real but conditional on representation design and classifier choice. \\

Signature-based representations outperform raw baselines when paired with suitable models, with Random Forest achieving the best overall performance. Time augmentation provides the most effective balance between temporal expressiveness and computational tractability, while deeper truncation levels capture informative higher-order interactions, particularly for nonparametric models. In contrast, MLP performance is more sensitive to high-dimensional representations. \\

Beyond empirical performance, this study highlights that representation design is a central component of time-series modelling. Improvements in HAR are better explained by how temporal structure is encoded rather than by model complexity alone. \\

By providing a fully specified and reproducible framework, this work enables transparent comparison and establishes a foundation for future research on structured temporal representations in HAR and related domains. 

\subsection{Perspectives}

Several directions emerge from this study. First, future work should evaluate signature-based representations across a broader range of datasets, including real-world and clinical HAR settings, to assess generalisability beyond the UCI HAR benchmark. In particular, datasets with variable-length sequences or less structured segmentation would provide a more rigorous test of representation robustness \parencite{banos2014windowsize,hammerla2016}. \\

Second, integrating signature features with deep learning architectures represents a promising avenue \parencite{deepsignaturetransforms}. Hybrid approaches that combine explicit signature representations with convolutional or recurrent networks may capture both structured interactions and hierarchical temporal patterns \parencite{deepsignaturetransforms}. In this context, deep signature methods, which embed signature transforms within neural network architectures, offer a particularly compelling direction by enabling end-to-end learning of both representation and classification \parencite{deepsignaturetransforms}. \\

Third, further work is needed to address the computational challenges associated with high truncation depths. Techniques such as log-signatures, feature selection, or regularisation may reduce dimensionality while preserving informative structure \parencite{kidger2021, dimensionreduction}. \\

Finally, exploring adaptive or learned embedding strategies, rather than fixed designs such as time augmentation or lead--lag, may provide more flexible and data-driven representations, extending the applicability of signature-based methods in time-series classification \parencite{morrill2021generalisedsignaturemethodmultivariate, deepsignaturetransforms}. \\



\section*{Declaration of Competing Interests}

The authors confirm that there are no financial or personal affiliations that may have biased the research presented in this paper.

\section*{Data Availability}

The dataset used in this study is publicly available at the UCI Machine Learning Repository\footnote{\href{http://archive.ics.uci.edu/dataset/240/human+activity+recognition+using+smartphones}{Human Activity Recognition Using Smartphones dataset}}. No additional data were generated or analysed during the current study.

\section*{Acknowledgements}

This work was supported by the Consulat Général de France à Jérusalem and the Communauté d'Agglomération Pau Béarn Pyrénées (CAPBP).

\printbibliography

\begin{appendix}
\section{Dataset and data representation}
\label{imple}

This section describes the dataset used in our experiments and the different data representations considered in the analysis. We first introduce the UCI HAR dataset and its input signals, and then describe the flattened, time-augmented, and lead--lag representations used throughout the experiments. 
\subsection{Dataset}

Experiments use the UCI Human Activity Recognition (HAR) dataset \parencite{anguita2013public,ucihar2012}, comprising recordings from 30 participants performing six activities: walking, walking upstairs, walking downstairs, sitting, standing, and laying. Data were acquired via a smartphone-mounted accelerometer and gyroscope at 50 Hz and segmented into 128-step windows (2.56 seconds) with 50\% overlap. Further details are provided in \textcite{anguita2013public}; the dataset is publicly available at the UCI Machine Learning Repository\footnote{\href{http://archive.ics.uci.edu/dataset/240/human+activity+recognition+using+smartphones}{Human Activity Recognition Using Smartphones dataset}}. \\

The analysis considers raw inertial signals from body acceleration and gyroscope measurements along the $x$, $y$, and $z$ axes, forming six-channel inputs of shape $(128,6)$. Additional experiments use only the three acceleration channels, yielding inputs of shape $(128,3)$. \\

The predefined train--test split is preserved (7,352 training and 2,947 test windows), with the test set treated as a strict hold-out and excluded from all model development and validation. \\

The dataset also provides 561 handcrafted time- and frequency-domain features, used solely for baseline reproduction. Primary analyses compare raw and signature-based representations within a unified, leakage-aware framework. 

\subsection{Data Representation}

We use three different representations of the UCI HAR dataset: time-augmented, lead--lag and flattened as defined in Section \ref{embed} and Section \ref{flat}.\\

Each observation is a multivariate time series in $\R^d$ defined over a time-scale $\T$ with $N=127$ and $d=6$ corresponding to the three-axis body acceleration and the three-axis gyroscope measurements. \\

Each data stream is flattened to a 768-dimensional vector. \\

A reduced setting uses only acceleration signals, yielding $(128,3)$ inputs flattened to 384 dimensions. \\

To enable path-based representations, a normalized time variable $t \in [0,1]$ is appended as an additional channel \parencite{FERMANIAN2021107148}, yielding $(128,7)$ and $(128,4)$ representations for the six- and three-channel settings, respectively. \\

A lead--lag transformation is then applied to the time-augmented paths to capture temporal dependencies and cross-channel interactions \parencite{Yang2022,10.1145/2640087.2644157}. For a $d$-dimensional path of length $T$, this produces a path of dimension $2d$ and length $2T$ (e.g., $(128,7) \rightarrow (256,14)$). \\

All representations are evaluated under identical experimental conditions to isolate their effect on classification performance. 

\section{Model Implementation Details}
\label{modelimple}

The MLP is implemented in PyTorch and operates on fixed-length inputs \parencite{Imambi2021}. The architecture consists of fully connected layers with batch normalisation and dropout, and outputs class probabilities over six activities. Training uses AdamW (learning rate $10^{-3}$, weight decay $10^{-4}$, batch size 256, maximum 50 epochs) with early stopping (patience 8) based on validation macro F1. We evaluate: (i) six-channel inputs with ReLU activation function, both with signature features and with raw flattened signals; and (ii) three-channel inputs with both ReLU and Tanh activation functions, again evaluated with and without signature features. \\

The RF classifier is implemented using \texttt{RandomForestClassifier} with warm-start training \parencite{10.5555/1953048.2078195}. The ensemble grows from 1 to a maximum of 50 trees, with early stopping after 8 non-improving iterations. Both Gini impurity and entropy criteria are evaluated. We evaluate: (i) six-channel inputs with signature features under both splitting criteria; (ii) six-channel raw flattened inputs evaluated with both Gini and entropy \textcite{breiman1984classification}; and (iii) an extended analysis using time-augmented signature features with truncation depths $m \in \{2,\dots,6\}$ under the entropy criterion. \\

All models are evaluated under identical data splits, preprocessing, and validation procedures. 

\subsection{Training Protocol}

All experiments were conducted under a fixed, leakage-free protocol applied consistently across models and feature representations \textcite{jarrar2026protocol}.\\

We retained the original UCI HAR train--test split \parencite{anguita2013public} (7,352/2,947 samples), with the test set used exclusively for final evaluation. Model selection was performed using a subject-disjoint validation split (15\% of the training data) generated via \texttt{GroupShuffleSplit} \parencite{10.5555/1953048.2078195}, grouping by subject identifiers to prevent cross-subject leakage \parencite{dehghani2019subjectcv,tello2024toogood}. The split was fixed with a random seed of 2025 and reused across all experiments.\\

All features were standardised using \texttt{StandardScaler} fitted on the training subset only \parencite{10.5555/1953048.2078195}. Model selection was based on validation macro F1 score. For the MLP, early stopping with a patience of 8 epochs was applied; for Random Forest, trees were added incrementally with stopping after 8 non-improving iterations.\\

All experiments used identical data partitions, preprocessing, and evaluation metrics. Stochastic components were controlled via a fixed random seed (2025), and all intermediate artefacts were saved to ensure full reproducibility. 

\subsection{Evaluation Metrics}

Model performance is evaluated using complementary metrics to provide a comprehensive assessment of classification accuracy, class-wise behaviour, and uncertainty.\\

Accuracy is defined as the proportion of correctly classified samples \parencite{FAWCETT2006861}:
\begin{equation}
\text{Accuracy}
=
\frac{1}{n}
\sum_{i=1}^{n}
\mathbb{I}(\hat{y}_i = y_i),
\label{eq:accuracy}
\end{equation}
where $n$ denotes the number of evaluated samples, and
$\hat{y}_i$ and $y_i$ denote the predicted and true labels,
respectively.\\

To ensure balanced evaluation across classes, the macro-averaged F1 score
is used as the primary metric for model selection
\parencite{manning2008introduction}. Let $C$ denote the total number of
activity classes and let $c\in\{1,\ldots,C\}$ denote the index of a
particular class. In this study, $C=6$. For each class $c$, precision and
recall are defined as
\begin{equation}
P_c = \frac{TP_c}{TP_c + FP_c},
\qquad
R_c = \frac{TP_c}{TP_c + FN_c},
\end{equation}
where $TP_c$, $FP_c$, and $FN_c$ denote the numbers of true positives,
false positives, and false negatives for class $c$, respectively.

The F1 score for class $c$ is then defined as
\begin{equation}
F1_c = \frac{2P_cR_c}{P_c + R_c}.
\label{eq:f1_class}
\end{equation}

The macro-averaged F1 score is obtained by averaging the class-specific
F1 scores:
\begin{equation}
\text{Macro-F1}
=
\frac{1}{C}\sum_{c=1}^{C} F1_c,
\label{eq:macro_f1}
\end{equation}
where $C=6$ is the number of activity classes. Macro-F1 assigns equal
weight to each class, independently of its number of observations.\\

Confusion matrices are computed on the test set to analyse class-specific performance, with both raw counts and row-normalised values reported. Additionally, classification reports provide per-class precision, recall, and F1 scores for detailed evaluation.\\

For interpretability, the number and proportion of correctly and incorrectly classified samples are also reported. To quantify uncertainty, 95\% bootstrap confidence intervals (1000 resamples) are computed for both accuracy and macro F1 \parencite{Efron1995AnIT}.\\

Together, these metrics provide a comprehensive and interpretable evaluation of model performance.\\

For the baseline reproduction study (Section 3.3), performance is assessed using mean accuracy from 10-fold stratified cross-validation, with emphasis on the proportion of correctly classified instances, consistent with the original study \parencite{IJAIN54|to_array:0}. 

\subsection{Reproducibility Framework}

This study adopts a fully reproducible framework in which all components—data preprocessing, feature construction, model training, evaluation, and reporting—are explicitly defined and publicly accessible.\\

A pre-specified protocol, defined prior to analysis and available on the Open Science Framework (OSF)\footnote{\url{https://osf.io/9k4rc}}, documents all methodological decisions. All experiments are implemented in Python and released in a public GitHub repository\footnote{\url{https://github.com/pwd529/har-signature-ML-reproducibility}} containing complete pipelines.\\

To support transparency and reuse, a model card for the best-performing model is provided via Hugging Face\footnote{\url{https://huggingface.co/KamalJarrar/har-signature-rf-classic-time-depth6}} \parencite{10.1145/3287560.3287596}, and a Docker container\footnote{\url{https://hub.docker.com/r/pwd529/har-signature-best-model}} ensures full computational reproducibility. Deterministic execution is enforced through fixed random seeds, and all configurations and outputs are systematically stored to ensure traceability.\\

Together, these components establish a transparent, verifiable, and reproducible research framework. 

\section{About Ismi et al. approach using K-means-Based Feature Reduction}
\label{ismi}

To benchmark against prior work, we reproduce the K-means-based feature reduction and Naive Bayes classification pipeline proposed in \parencite{IJAIN54|to_array:0}. The approach operates on the 561-dimensional handcrafted features provided with the UCI HAR dataset \parencite{anguita2013public}.\\

Feature reduction is performed by applying K-means clustering to the transposed feature matrix, treating each feature as an observation. For $k \in \{10, 20, 30, 40, 50, 60, 70\}$, features are grouped into $k$ clusters, and cluster-wise averaging yields a reduced representation of dimension $(N, k)$. The study reports a baseline accuracy of 76.77\% using all 561 features and a maximum of 83.35\% after feature reduction (achieved at $k=70$) \parencite{IJAIN54|to_array:0}.\\

As the original implementation in \parencite{IJAIN54|to_array:0} lacks full specification (e.g., preprocessing and parameter settings), this implementation follows the published description and documents all assumptions explicitly to ensure transparency and reproducibility.\\

In our reproduction, K-means is implemented with $n\_clusters = k$, random initialisation, \texttt{n\_init=1}, \texttt{max\_iter=5}, and \texttt{random\_state=42}. \\

The reported results in \parencite{IJAIN54|to_array:0} are compared against those from our reproduction and the proposed methodology. \\

\end{appendix}
\end{document}